\documentclass[runningheads]{llncs}
\usepackage[T1]{fontenc}
\usepackage{graphicx}
\usepackage{amsmath}
\usepackage{bm}
\DeclareMathOperator*{\argmax}{arg\,max}
\usepackage[linesnumbered, lined,boxed,commentsnumbered, ruled, vlined]{algorithm2e}
\usepackage[T1]{fontenc}    
\usepackage{hyperref}       
\usepackage{url}            
\usepackage{graphicx}    
\usepackage{float}
\usepackage{booktabs}  
\usepackage{amsfonts} 
\usepackage{dsfont}
\usepackage{subfig} 
\usepackage{graphicx}
\usepackage{amsthm}
\usepackage{amssymb}
\usepackage[table,xcdraw]{xcolor}
\usepackage{nicefrac}       
\usepackage{microtype}      
\usepackage{lipsum}

\usepackage[most]{tcolorbox}

\definecolor{olive}{rgb}{0.6, 0.6, 0.2}
\definecolor{sand}{rgb}{0.8666666666666667, 0.8, 0.4666666666666667}
\definecolor{wine}{rgb}{0.5333333333333333, 0.13333333333333333, 0.3333333333333333}
\definecolor{deblue}{RGB}{11,132,147}
\definecolor{ocra}{RGB}{204, 119, 34}

\newcommand{\textbfr}[1]{\textbf{\textcolor{red}{#1}}}
\newcommand{\textbfb}[1]{\textbf{\textcolor{blue}{#1}}}

\newtcolorbox{CatchyBox}[2][]{
    lower separated=false,
    colback=white!90!blue!90!ocra,
    colframe=white, fonttitle=\bfseries,
    colbacktitle=white!70!blue!90!ocra,
    coltitle=black,
    enhanced,
    attach boxed title to top left={xshift=.02\linewidth,yshift=-4mm},
    title=#2,#1}

\usepackage{pifont}
\newcommand{\cmark}{\ding{51}}%
\newcommand{\xmark}{\ding{55}}%

\begin{document}

\title{Spectral Ranking with Covariates}


\author{Siu Lun Chau \and Mihai Cucuringu  \and Dino Sejdinovic}
\authorrunning{Chau et al.}
%
\institute{Department of Statistics, University of Oxford, 29 St Giles' Oxford OX1, UK \\
\email{\{siu.chau, mihai.cucuringu, dino.sejdinovic\}@stats.ox.ac.uk}\\
}

\maketitle              
\begin{abstract}
We consider spectral approaches to the problem of ranking $n$ players given their incomplete and noisy pairwise comparisons, but revisit this classical problem in light of player covariate information. We propose three spectral ranking methods that incorporate player covariates and are based on \textit{seriation}, \textit{low-rank structure} assumption and \textit{canonical correlation}, respectively. Extensive numerical simulations on both synthetic and real-world data sets demonstrated that our proposed methods compare favorably to existing state-of-the-art covariate-based ranking algorithms.
\end{abstract}

\def\NN{{\mathbb N}}    
\def\ZZ{{\mathbb Z}}     
\def\RR{{\mathbb R}}
\def\QQ{{\mathbb Q}}    
\def\CC{{\mathbb C}}    
\def\HH{{\mathbb H}}    
\def\PP{{\mathbb P}}     
\def\KK{{\mathbb K}}     
\def\EE{{\mathbb E}}    
\def\11{{\mathbf 1}}    
\def\AA{{\mathbb A}}

\def\cA{{\mathcal A}}  \def\cG{{\mathcal G}} \def\cM{{\mathcal M}} \def\cS{{\mathcal S}} \def\cB{{\mathcal B}}  \def\cH{{\mathcal H}} \def\cN{{\mathcal N}} \def\cT{{\mathcal T}} \def\cC{{\mathcal C}}  \def\cI{{\mathcal I}} \def\cO{{\mathcal O}} \def\cU{{\mathcal U}} \def\cD{{\mathcal D}}  \def\cJ{{\mathcal J}} \def\cP{{\mathcal P}} \def\cV{{\mathcal V}} \def\cE{{\mathcal E}}  \def\cK{{\mathcal K}} \def\cQ{{\mathcal Q}} \def\cW{{\mathcal W}} \def\cF{{\mathcal F}}  \def\cL{{\mathcal L}} \def\cR{{\mathcal R}} \def\cX{{\mathcal X}} \def\cY{{\mathcal Y}}  \def\cZ{{\mathcal Z}}


\def\mfA{{\mathfrak A}} \def\mfA{{\mathfrak P}} \def\mfS{{\mathfrak S}}\def\mfZ{{\mathfrak Z}} \def\mfM{{\mathfrak M}} \def\mfQ{{\mathfrak Q}} \def\mfE{{\mathfrak E}} \def\mfL{{\mathfrak L}} \def\mfW{{\mathfrak W}} \def\mfR{{\mathfrak R}} \def\mfK{{\mathfrak K}} \def\mfX{{\mathfrak X}} \def\mfT{{\mathfrak T}} \def\mfJ{{\mathfrak J}} \def\mfC{{\mathfrak C}} \def\mfY{{\mathfrak Y}} \def\mfH{{\mathfrak H}} \def\mfV{{\mathfrak V}}\def\mfU{{\mathfrak U}}\def\mfG{{\mathfrak G}} \def\mfB{{\mathfrak B}} \def\mfI{{\mathfrak I}} \def\mfF{{\mathfrak F}} \def\mfN{{\mathfrak N}} \def\mfO{{\mathfrak O}} \def\mfD{{\mathfrak D}} 

\def\mfa{{\mathfrak a}} \def\mfp{{\mathfrak p}} \def\mfs{{\mathfrak s}}  \def\mfz{{\mathfrak z}} \def\mfm{{\mathfrak m}} \def\mfq{{\mathfrak q}}  \def\mfe{{\mathfrak e}} \def\mfl{{\mathfrak l}} \def\mfw{{\mathfrak w}} \def\mfr{{\mathfrak r}} \def\mfk{{\mathfrak k}} \def\mfx{{\mathfrak x}} \def\mft{{\mathfrak t}} \def\mfj{{\mathfrak j}} \def\mfc{{\mathfrak c}} \def\mfy{{\mathfrak y}} \def\mfh{{\mathfrak h}} \def\mfv{{\mathfrak v}} \def\mfu{{\mathfrak u}} \def\mfg{{\mathfrak g}} \def\mfb{{\mathfrak b}} \def\mfi{{\mathfrak i}} \def\mff{{\mathfrak f}} \def\mfn{{\mathfrak n}} \def\mfo{{\mathfrak o}} \def\mfd{{\mathfrak d}}
\global\long\def\bone{{\bf 1}}

\def\boldx{{\boldsymbol x}} \def\boldt{{\boldsymbol t}}
\def\bfx{{\bf x}} \def\bfy{{\bf y}} \def\bfz{{\bf z}} \def\bfw{{\bf w}}
\def\bfk{{\bf k}} \def\bfK{{\bf K}} \def\bfell{{\bf \ell}}
\def\bfL{{\bf L}} \def\bfQ{{\bf Q}} \def\bfA{{\bf A}}
\def\bfPhi{{\bf \Phi}} \def\bfPsi{{\bf \Psi}}
\def\boldupsilon{\boldsymbol{\Upsilon}}
\def\bfeta{\boldsymbol{\eta}} \def\bfSigma{\boldsymbol{\Sigma}}

\def\bfr{{\bf r}}
\def\bfa{{\bf a}}
\def\bfb{{\bf b}}
\def\bfV{{\bf V}}
\def\bfz{{\bf z}}
\def\bfq{{\bf q}}
\def\bfH{{\bf H}}
\def\bfC{{\bf C}}
\def\bfG{{\bf G}}
\def\bfD{{\bf D}}
\def\bfS{{\bf S}}
\def\bfs{{\bf s}}
\def\bfZ{{\bf Z}}
\def\bfX{{\bf X}}
\def\bfB{{\bf B}}
\def\bfI{{\bf I}}
\def\bfSigma{{\bm{\Sigma}}}
\def\bfbeta{{\bm{\beta}}}
\def\bfalpha{{\bm{\alpha}}}
\def\bfgamma{{\bm{\gamma}}}
\def\bfPsi{{\bm{\Psi}}}

\def\rmC{{\mathrm C}}
\def\rmD{{\mathrm D}}
\def\rmc{{\mathrm c}}
\def\rmd{{\mathrm d}}

\def\tilx{{\tilde{x}}}
\def\tilbfx{{\bf \tilde{x}}}

\def\balpha{\boldsymbol{\alpha}}



\newcommand{\indep}{\perp \!\!\!\!\!\; \perp}

    
\newcommand{\deffunction}[5]{
{#1}:
\left|
  \begin{array}{rcl}
    {#2} & \longrightarrow & {#3} \\
    {#4} & \longmapsto & {#5} \\
  \end{array}
\right.
}

\newcommand{\restrict}[2]{{#1}_{\mkern 2mu \vrule height 2.5ex\mkern2mu {#2}}}

\def\d{\,{\mathrm d}}

\def\tr{\operatorname{tr}}
\newcommand{\Range}[1]{\operatorname{Range}({#1})}
\newcommand{\Span}[1]{\operatorname{Span}\left\{{#1}\right\}}
\newcommand{\cSpan}[1]{\overline{\operatorname{Span}}\left\{{#1}\right\}}

\vspace{-0.3cm}
\section{Introduction}
\label{section: introduction}

We consider the classical problem of ranking $n$ players, given a set of pairwise comparisons, but revisiting this problem in light of available player covariate information. We assume ranking is represented by a vector $\bfr \in \RR^n$, which can be interpreted as a one-dimensional representation of players' underlying skills. In practice, such pairwise comparisons are often incomplete and inconsistent with respect to the underlying skill vector $\bfr$; therefore, the objective of a ranking problem is often to recover a total ordering of the players that is as consistent as possible with the available noisy and sparse information.



There exists a rich literature on ranking that can be dated back as early as the seminal work of Kendall and Smith~\cite{KendallSmith1940} in the 1940s. Classical examples such as the Bradley-Terry-Luce model~\cite{bradley1952rank}, Random Utility model~\cite{thurstone1927law.}, Mallows-Condorcet model~\cite{Mallows1957}, and Thurstone-Mosteller model~\cite{mosteller1951experimental} have inspired numerous developments of modern ranking algorithms~\cite{cattelan2012models}. These methods take a probabilistic approach, in which they treat ranking as the output generated from a probabilistic model, where maximum likelihood estimators can be built subsequently. 



Apart from probabilistic approaches, spectral methods that apply the theory of linear maps (in particular, eigenvalues and eigenvectors) to the pairwise comparison matrix or derivations of it, in order to extract rankings, is also a century-old~\cite{vigna2016spectral} idea that was made famous by Google's PageRank~\cite{page1999pagerank} algorithm. There are several other popular spectral ranking methods proposed in recent years as well. For example, Fogel et al.~\cite{serialRank} connected the problem of seriation~\cite{atkins1998spectral} to ranking and proposed \textsc{SerialRank} algorithm. Cucuringu et al.\cite{SVDRank}, on the other hand, proposed \textsc{SVD-Rank}, a simple SVD-based approach to recover rankings when cardinal comparisons are observed. Cucuringu also proposed \textsc{SyncRank}~\cite{syncRank} based on group synchronization using semidefinite programming  and spectral relaxations.

In many instances, it is of interest to investigate whether some player covariates affect the results of the comparisons. It is also natural to believe that incorporating informative covariates will lead to better ranking estimations. Besides improving estimation, we believe there are at least two more reasons why this might be useful in practice:
\begin{enumerate}
    \item Covariate-free ranking algorithms cannot take into account instances where new players are added to the comparison sets, without first observing new matches with existing players. This can be overcome by building a model of the ranking as a function of player covariates.
    \item Traditional ranking methods often break down if the sample complexity of the comparison graph does not scale as $O(n\log n)$, meaning the graph is disconnected with high probability~\cite{SVDRank}. By utilising the smoothness of rankings across covariate information, one is able to overcome this obstacle and infer meaningful rankings even in the cases of very sparse comparisons that render the graph disconnected.
\end{enumerate}

A number of extensions to incorporate covariates have been proposed for probabilistic approaches~\cite{chu2005preference,niranjan2017inductive,springall1973response}. In contrast, spectral ranking with covariates received much lesser attention. To the best of our knowledge, the recent work of Jain et al.~\cite{jain2020spectral} was the first to incorporate covariates into spectral ranking. However, unlike our approaches, their approach cannot be used to predict rankings on new players. 

To this end, we introduce a suite of spectral ranking algorithms incorporating covariate information: \textsc{C-SerialRank}, \textsc{SVDCovRank} and \textsc{CCRank}. Both \textsc{C-SerialRank} and \textsc{SVDCovRank} extend on existing spectral ranking techniques, while \textsc{CCRank} is a new spectral ranking approach motivated by analysing the canonical correlation between covariate information and match outcomes. All proposed methods appeal to a class of expressive non-parametric models based on reproducing kernel Hilbert spaces. Our contributions are summarised as follows:
\begin{enumerate}
    \item We extend \textsc{SerialRank} and \textsc{SVDRank} to incorporate player covariate information and propose a new spectral ranking algorithm \textsc{CCRank}. This method extracts rankings based on the canonical correlation between covariate and match information.
    \item We provide an extensive set of numerical experiments on simulated and real-world data, showcasing the utility of including covariate information as well as the competitiveness of the proposed methods with state-of-the-art covariate-based ranking algorithms.
\end{enumerate}

\paragraph{Outline.} The rest of the paper is organised as follows. Section \ref{section: related work} covers related work in the spectral ranking literature. Our proposed methodologies will be presented in Section \ref{section: proposed methodology}. Sections \ref{section: experiments} and \ref{section: discussion} cover experiments and concluding remarks.

\paragraph{Notation.} Scalars are denoted by lower case letters while vectors and matrices are denoted by bold lower and upper case letters, respectively. In general, there are $n$ players to rank, and we denote the $n\times n$ pairwise comparison matrix by $\bfC$. We use the notation $\succ$ to denote the direction of preference. For ordinal comparisons, $C_{i, j} = 1$ if $i \succ j$, and $-1$ otherwise. For simplicity, we assume no ties between players. For \textit{cardinal} comparisons, $C_{i,j} \in \RR$ is considered as a proxy or noisy evaluation of the skill offset $r_i - r_j$ as in \cite{syncRank,SVDRank}. We denote the player covariate information matrix as $\bfX\in\RR^{n\times d}$ and its subsequent kernel matrix as $\bfK \in \RR^{n \times n}$ with $K_{i,j}= k(\bfx_i, \bfx_j)$, where $\bfx_i$ is the $i^{\text{th}}$ row of $\bfX$ and $k$ the kernel. Given a skill vector $\bfr$, we define upsets as a pair of items for which the player with lower skill is preferred over the player with higher skill, i.e. there is an upset if the recovered ranking indicates $r_i > r_j$ but the input data has $C_{i, j} < 0.$. Finally, the ranking is inferred by sorting the entries of $r$, choosing either increasing or decreasing order to \textit{minimise} the number of upsets with respect to the observed $\bfC$.







\section{Background}
\label{section: related work} 

We give a brief review of both probabilistic and spectral ranking algorithms in this section. 


\subsection{Probabilistic Ranking.}
Probabilistic ranking methods typically use a generative model of how the pairwise comparison is conducted: given players' skill vector $\bfr$, also denoted as \textit{worths} or \textit{weights parameters}, one can model the match outcome probability based on the difference between the skills of the two players, i.e. $\PP(i \succ j) = F(r_i -  r_j)$, where $F$ is some cumulative distribution function (CDF). The Thurstone~\cite{thurstone1927law.} model uses a normal CDF for $F$ while the Bradley-Terry-Luce (\textsc{BT}) model uses a logistic CDF. This simple setup has led to various extensions. For example, Huang et al.~\cite{huang2006generalized} proposed a generalisation of Bradley-Terry models to tackle paired team comparisons instead of individuals comparisons. Chen et al.~\cite{chen2016} studied intransitivity in comparison data by extending ranking from a one-dimensional vector to a multi-dimensional representation. Bayesian approaches taking into account prior beliefs over ranking are developed in \cite{chu2005preference,caron2012efficient}.

Extending \textsc{BT} models to incorporate player covariates has been studied thoroughly in the past decades as well. Springall~\cite{springall1973response} proposes to describe the skill vector $\bfr$ as the linear combination of player covariates $\bfX$, i.e. $\bfr = \bfX \bfbeta$ for some regression coefficients $\bfbeta \in \RR^d$. Chu et al.~\cite{chu2005preference} later extended this formulation to a more flexible Gaussian Processes~\cite{williams2006gaussian} regression model as well. We will use these two extensions and the \textsc{BT} model as our baselines in our numerical experiments.

\subsection{Spectral Ranking.}
\label{sec: spectral rankings}
Spectral ranking is a family of fundamentally different approaches and the core object of interest are the matrices that represent pairwise relationships between the players. It often involves computing the leading eigenvector of the sparse comparison matrix, or similar matrix operators derived from it, which can be computed efficiently in contrast to the optimisation procedures inherent in the MLE~\cite{jain2020spectral}. The computational advantage stems from the fact that typical eigen-solvers can compute the leading eigenvectors of a sparse matrix in running time linear in the number of nonzero entries in the matrix (ie. edges of the comparison graph). We refer the reader to Vigna~\cite{vigna2016spectral} for a comprehensive review of the early history of spectral ranking. We now present two specific spectral ranking algorithms that our proposed methods are based on, and also discuss Jain et al.~\cite{jain2020spectral}'s spectral method that utilises covariates as a baseline in our numerical experiments.

\paragraph{\textsc{SerialRank.}} Fogel et al.~\cite{serialRank} proposed a seriation approach to ranking, building on \cite{atkins1998spectral}, where they defined similarities between players based on their pairwise comparisons. Given an ordinal pairwise comparison matrix $\bfC$, we define the similarity $S_{i,j}$ between player $i, j$ as the number of agreeing comparisons between $i, j$ with other players, i.e.
\begin{align}
\small
\label{eq: serial S}
    S_{i, j} = \sum_{k=1}^n \left(\frac{1+C_{i,k}C_{j,k}}{2}\right), 
\end{align}
which can be expressed as $\bfS = \frac{1}{2}(n\bone\bone^\top + \bfC\bfC^\top)$, where $\bone \in \RR^n$ denotes the all-one vector. The optimal ordering in this seriation process is then recovered by extracting the \textit{Fiedler vector} of $\bfS$, which is the eigenvector corresponding to the second smallest eigenvalue of the graph Laplacian matrix $\cL(\bfS) = \operatorname{diag}(\bfS\bone) - \bfS$. This corresponds to the following optimisation
\begin{equation}
    \bfr_* = \arg\min_\bfr \left(\bfr^{\top}\mathcal{L}(\bfS)\bfr\right) \quad \text{s.t  }\quad \bfr^{\top}\bfr = 1,\quad \bfr^{\top}\bone = 0.
\end{equation}
$\bfr_*$ can be interpreted as the smoothest non-trivial graph signal with respect to the similarity graph $\bfS$~\cite{shuman2013emerging}. 






\paragraph{\textsc{SVDRank}.} Cucuringu et al.~\cite{syncRank,SVDRank} considered a simple alternative spectral method, arising from models wherein the pairwise comparisons are modelled as noisy entries of a rank 2 skew-symmetric matrix $\bfC = \bfr \bone^\top - \bone \bfr^\top$, or a proxy of it. This leads to the following optimisation

\begin{equation}
\bfr_* = \arg\max_\bfr \left(\bfr^{\top}\bfC\bfC^{\top}\bfr\right),\quad  \text{s.t. }\quad \bfr^{\top}\bfr=1, \quad \bfr^{\top}\bone=0.\label{eq: SVD-Rank-orth}
\end{equation}
One computes the top two singular vectors of $\bfC$ and extracts the ranking induced by ordering their entries to minimise upsets. 




\paragraph{\textsc{Rank-Centrality}.} Negahban et al.~\cite{Negahban_RankCentrality_2017} proposed to estimate player rankings from the stationary distribution of a certain random walk on the graph of players, where edges encode the probabilities of pairwise comparison outcomes. This model was first designed to combine \textbf{multiple} \textit{ordinal} pairwise comparisons made on the same pairs of players under the context of \textit{Rank Aggregation}~\cite{li2019comparative}. Cucuringu~\cite{sync} later combined this approach with \textsc{SerialRank} to handle the single observation setting. Given the similarity matrix $\bfS$ computed using Eq~\eqref{eq: serial S}, the key idea behind \textsc{Rank-Centrality} is to construct a transition probability matrix $\bfA$ where $A_{i, j} = 1 - \frac{S_{i,j}}{2n}$ if $C_{i,j} > 0$, $\frac{S_i,j}{2n}$ if $C_{i,j} < 0$, and $0$ otherwise. One then extracts the leading top left-eigenvector from $\bfA$ as the ranking. Jain et al.~\cite{jain2020spectral} later extended this to incorporate covariates by setting $\hat{\bfA} = \bfA\bfV$ where $\bfV$ is some row-stochastic player similarity matrix, i.e., a scaled kernel matrix. Jain et al. called this method the \textsc{Regularised Rank-Centrality} algorithm. We note that this method cannot predict the ranking of unseen players because no explicit functional relationships between $\bfr$ and player covariates $\bfX$ are modelled.

\section{Proposed methods}
\label{section: proposed methodology}

In this section, we introduce three lines of spectral ranking algorithms, each having a different flavour in combining the two sources of data: pairwise comparisons and covariate information. \textsc{C-SerialRank} controls the information covariates contribute via a simple hyperparameter (that can be selected based on cross-validation), and therefore can recover a robust ranking even when features are noisy. On the other hand, \textsc{SVDCovRank} connects $\bfr$ with covariates via a functional model, allowing us to predict the rank of previously unseen players while no comparison with existing players is needed. Finally, \textsc{CCRank} considers the optimal embedding of both match outcomes and covariate information that maximises their canonical correlation and extracts ranking from those embeddings. 

\subsection{\textsc{C-SerialRank}}

Recall in \textsc{SerialRank}, ranking is extracted from the Fiedler vector of the graph Laplacian matrix $\cL(\bfS)$, where $\bfS = \frac{1}{2}(n\bone\bone^\top + \bfC\bfC^\top)$ is the player similarity matrix induced by their match outcomes. However, this is not the only source of player similarity. If we also have access to the covariate matrix $\bfX$, we can define another source of similarity through the Gram matrix $\bfK$, with $K_{i, j} = k(\bfx_i, \bfx_j)$ for a given kernel $k$. We propose to merge this extra source of similarity with $S$ via a simple linear combination, i.e. $\hat{\bfS} = \bfS + \lambda \bfK$ with $\lambda$ controlling the tradeoff between the two sources of data. We call this method \textsc{C-SerialRank}, where $\textsc{C}$ stands for covariates. The ranking can then be recovered by computing the Fiedler vector from the new graph Laplacian matrix $\cL(\hat{\bfS})$
\begin{equation}
\label{eq:constrained_SR}
    \bfr_* = \arg\min_{\bfr} \bfr^T\big(\mathcal{L}(\bfS + \lambda \bfK) \big)\bfr \quad \text{s.t }\quad \bfr^\top\bfr = 1\quad \bfr^\top\bone = 0.
\end{equation}
We summarise the method in Algorithm~\ref{algo: CSerialRank}. The tuning parameter $\lambda$ controls the contribution of the covariates -- intuitively, it should be high for covariates that are informative about match outcomes, and low for less informative covariates. 

\begin{algorithm}
\DontPrintSemicolon
\SetAlgoLined
\SetKwInOut{Input}{Require}
\Input{A set of pairwise comparisons $\bfC_{i,j} \in \{-1, 0, +1\}$ or $\RR$, kernel matrix $\bfK$, regularisation parameter $\lambda$.}
{Compute similarity matrix $\hat{\bfS} = \frac{1}{2}(n\bone\bone^T + \bfC\bfC^T) + \lambda \bfK$.} \;
{Compute the associated graph Laplacian matrix $\cL(\hat{\bfS}) = \operatorname{diag}(\hat{\bfS}\bone) - \hat{
\bfS}$}. \;
{Compute the Fiedler vector of $\bfS$ by extracting the eigenvector corresponding to the second smallest eigenvalue of $\cL(\hat{\bfS})$}.\;
{Output the ranking induced by sorting the Fiedler vector of $\hat{\bfS}$, with the global ordering chosen to minimise upsets with respect to $\bfC$.}
\caption{\textsc{C-SerialRank}}
\label{algo: CSerialRank}
\end{algorithm}

\subsection{\textsc{SVDCovRank}}
\textsc{C-SerialRank} provides a simple method to incorporate player covariates into ranking. However, similar to the \textsc{Regularised Rank-Centrality} algorithm, the downside of \textsc{C-SerialRank} is that one cannot perform predictions on unseen players based on available covariate information. To this end, we introduce \textsc{SVDCovRank}, an SVD based ranking algorithm that connects covariates to $\bfr$ using a functional model.

Recall that when solving for $\textsc{SVDRank}$, one converts Eq~\eqref{eq: SVD-Rank-orth} to a standard eigenvector problem by projecting  $\bfr$ onto $\left\{ \bone\right\} ^{\perp}$. This can be done by setting $\bfr = \bfZ\bfq$, where $\bfZ\in\mathbb{R}^{n\times\left(n-1\right)}$ is a matrix whose columns form an orthonormal basis of $\left\{ \bone\right\} ^{\perp}$. As a result, Eq~\eqref{eq: SVD-Rank-orth} can be rewritten as
\begin{equation}
\bfq_* = \arg\max_{\bfq}\left(\bfq^{\top}\bfZ^{\top}\bfC\bfC^{\top}\bfZ\bfq\right),\quad\text{s.t. }\bfq^{\top}\bfq=1.\label{eq: SVD-Rank-orth-q}
\end{equation}
By construction, $\bfr_* = \bfZ\bfq_*$ is orthogonal to $\bone$ where $\bfq_*$ is the top eigenvector of $\bfZ^{\top}\bfC\bfC^{\top}\bfZ$. Now consider the scenario where we also have access to player covariates $\bfX$, and we assume the final ranking implied by $\bfr$ varies smoothly as a linear function of these covariates, e.g. $\bfr = \bfX\bfbeta$. Following the same line of thought as above, one could set $\bfX\bfbeta = \bfZ\bfq$, and substitute $\bfq = \bfZ^\top\bfX\bfbeta$ into Eq~\eqref{eq: SVD-Rank-orth-q} to arrive at
\begin{equation}
    \bfbeta_* = \arg\max_\bfbeta \left(\bfbeta^\top\bfX^\top \bfH \bfC\bfC^\top \bfH\bfX\bfbeta \right), \quad \text{s.t. }\bfbeta^\top\bfX\bfH\bfX\bfbeta = 1,  \label{eq: SVD-Rank-orth-beta}
\end{equation}
where $\bfH = \bfZ\bfZ^\top$ is the centering matrix. To solve for $\bfbeta$, we further decompose our feature covariance matrix $\bfX^\top\bfH\bfX$ into $\bfL\bfL^\top$ via a Choleskey decomposition or SVD. By setting $\bfbeta = \bfL^{-\top}\bfgamma$, we again recover the standard eigenvector problem
\begin{equation}
    \bfgamma_* = \arg\max_{\bfgamma} \left(\bfgamma^\top \bfPsi \bfgamma\right), \quad \text{s.t. }\bfgamma^\top \bfgamma = 1, \label{eq: SVD-Rank-orth-gamma}
\end{equation}
with $\bfPsi = \bfL^{-1}\bfX^\top \bfH\bfC\bfC^\top \bfH\bfX \bfL^{-
\top}$. Finally, the optimal ranking $\bfr$ is recovered by setting $\bfr_* = \bfH\bfX\bfL^{-\top}\bfgamma_{*}$. We summarise the procedure in Algorithm \ref{algo: SVDCovRank}.


Furthermore, if there is reason to believe the skill vector $\bfr$ relates to the covariates non-linearly, it is straightforward to kernelise the procedure by setting $\bfr = \bfK\bfalpha$, where $\bfK$ is the kernel matrix of player covariates and $\bfalpha \in \RR^n$ some dual weights, and obtain a non-parametric variation of \textsc{SVDCovRank}. Similarly by setting $\bfK\bfalpha = \bfZ^\top \bfq$, we can express Eq~\eqref{eq: SVD-Rank-orth-q} as
\begin{equation}
    \bfalpha_* = \arg\max_{\bfalpha} \left(\bfalpha^\top \bfK\bfH\bfC\bfC^\top \bfH\bfK\bfalpha\right), \quad \text{s.t. }\bfalpha^\top \bfK\bfH\bfK\bfalpha = 1.\label{eq: SVD-Rank-orth-alpha}
\end{equation}
The rest follows in analogy to the linear case and the complete method is given in Algorithm \ref{algo: SVDCovRank}.

\begin{algorithm}[t]
\SetKwInOut{Input}{Require}
\DontPrintSemicolon
\Input{ A set of pairwise comparisons $\bfC_{i, j} \in \{-1, 0, 1\}$  or $\mathbb{R}$, covariate matrix $\bfX$ and kernel function $k$.}
\If{kernelise}
{
Compute the kernel matrix $\bfK$ with $K_{i,j} = k(\bfx_i, \bfx_j)$ and set $\bfB = \bfK$
}
\Else
{
Set $\bfB=\bfX$.
}
{Compute $\bfPsi = \bfL^{-1}\bfB^{\top}\bfH\bfC\bfC^{\top}\bfH\bfB \bfL^{-\top}$, where $\bfL$ is the Cholesky decomposition of $\bfB^\top \bfH \bfB$.} \;
{Compute the top eigenvector $\bfgamma$ of $\bfPsi$ and set $\bfr = \bfH\bfB \bfL^{-\top}\bfgamma$.}\;
{Sort $\bfr$ to minimise the number of upsets with respect to $\bfC$.} \;
\If{kernelise}
{Return the skill vector $\bfr$ and skill function $r(\bfx) = k(\bfx, \bfX) \bfL^\top\bfgamma$ where $k(\bfx, \bfX) = [k(\bfx, \bfx_1), ..., k(\bfx, \bfx_n)]$.}
\Else
{Return the skill vector $\bfr$ and skill function $r(\bfx) = \bfx^\top \bfL^\top\bfgamma$.}
\caption{\textsc{SVDCovRank}}
\label{algo: SVDCovRank}
\end{algorithm}

\subsection{\textsc{CCRank}}
Besides aggregating match and covariate information via a linear combination or by imposing a regression structure between $\bfr$ and $\bfX$, we propose to study the Canonical Correlation between match and covariate information to derive the ranking. We call the method \textsc{CCRank}. Consider the similarity matrix $\bfS = \frac{1}{2}(n\bone\bone^\top + \bfC\bfC^\top)$ from Section~\ref{sec: spectral rankings} and player covariate matrix $\bfX$. As each row of $\bfS$ and $\bfX$ describe the same player with two distinct sources of information, it is natural to define ranking as the ``most agreeing'' projections between $\bfS$ and $\bfX$. This coincides with the classical Canonical Correlation Analysis (CCA) problem.



CCA is a classical technique developed by H. Hotelling~\cite{hotelling1992relations} to study the linear relationship between two multidimensional sets of variables. Assume both matrix $\bfS$ and $\bfX$ are full rank, CCA can be defined as the problem of finding the optimal projection $\bfz_S = \bfS\bfa$ and $\bfz_X = \bfX\bfb$, with $\bfa\in \RR^n$ and $\bfb\in\RR^d$, such that $\bfz_S$ and $\bfz_X$ are maximally correlated, i.e.
\begin{align}
\small
    \hat{\bfa}, \hat{\bfb} = \argmax_{\bfa, \bfb} \frac{\bfz_S^\top \bfz_X}{||\bfz_S||||\bfz_X||} = \argmax_{\bfa, \bfb} \frac{\bfa^\top \bfSigma_{SX}\bfb}{\sqrt{\bfa^\top \bfSigma_{SS}\bfa \bfb^\top \bfSigma_{XX}\bfb}},
\label{eq: CCA}
\end{align}
where $\bfSigma_{SX}=\bfS\bfX$ is an estimate of the cross-covariance matrix between $\bfS$ and $\bfX$. Eq~\eqref{eq: CCA} is equivalent to the following constrained optimisation problem
\begin{align}
\small
    \hat{\bfa}, \hat{\bfb} = \argmax_{\bfa, \bfb} = \bfa^\top \bfSigma_{SX}\bfb \quad \text{s.t. }
    \begin{cases}
    \bfa^\top \bfSigma_{SS}\bfa = 1\\
    \bfb^\top \bfSigma_{XX}\bfb = 1.
    \end{cases}
\end{align}
The solution to this problem can then be obtained by solving a generalised eigenvector problem as detailed in Algorithm~\ref{algo: CCARank}. We recover ranking by setting $\bfr$ to be $\bfz_S$ or $\bfz_X$ depending on which embedding better minimises the upsets with respect to $\bfC$. However, note that $\bfz_S = \bfS \bfa$ cannot be used to predict rankings on unseen players by construction.

Similar to \textsc{SVDCovRank}, if we believe the covariates relate to the optimal ranking non-linearly, we could kernelise the procedure and seek projections belonging to RKHSs. Denote $\ell$ as the kernel on $\bfS$ and $\cH_\ell$ the corresponding RKHS. The objective of Kernel CCA~\cite{fukumizu2007statistical} can be expressed using cross-covariance operators, which for simplicity, we use the same notation as the corresponding cross-covariance matrices
\begin{align}
\small
    \hat{f}, \hat{g} = \argmax_{f \in \mathcal{H}_k, g\in \mathcal{H}_\ell} \langle f,\bfSigma_{X S} g \rangle_{\mathcal{H}_\ell} \quad
     \text{s.t } \begin{cases}
    \langle f,\bfSigma_{XX} f \rangle_{\cH_k} = 1 \\
    \langle g, \bfSigma_{SS} g \rangle_{\cH_\ell} = 1 .
    \end{cases}
\end{align}
The rest of the algorithm is outlined in Algorithm~\ref{algo: CCARank}. Different from $\textsc{SVDCovrank}$ which postulates a low-rank structure on $\bfC$, $\textsc{CCRank}$ does not pose any structure over the match matrix, and thus is more robust to model misspecification. This is illustrated later in our experiments.


\IncMargin{1em}
\begin{algorithm}[t]
\SetKwInOut{Input}{Require}
\DontPrintSemicolon
\Input{ A set of pairwise comparisons $C_{i, j} \in \{-1, 0, 1\}$  or $\mathbb{R}$, covariate matrix $\bfX$, kernel functions $k, g$, centering matrix $\bfH$, identity matrix $\bfI_n$ and small perturbation $\epsilon$}
{Compute $\bfS = \frac{1}{2}(n\bone\bone^\top + \bfC\bfC^\top)$} \;
\If{kernelise}
{
{Compute the kernel matrix $\bfK$ with $K_{i,j} = k(\bfx_i, \bfx_j)$ and $\bfG$ with $G_{i,j} = g(\bfs_i, \bfs_j)$ where $\bfs_i,\bfs_j$ corresponds to row $i,j$ of $\bfS$.} \;
{Compute $\tilde{\bfK} = \bfH \bfK \bfH$ and $\tilde{\bfG} = \bfH \bfG \bf H$} \;
{Compute $\bfSigma_{XS} = n^{-1}\tilde{\bfK}\tilde{\bfG}, \bfSigma_{XX} = n^{-1}\tilde{\bfK}(\tilde{\bfK} + \epsilon \bfI_n)$ and $\bfSigma_{SS} = n^{-1}\tilde{\bfG}(\tilde{\bfG} + \epsilon \bfI_n)$}\;
}
\Else
{
Compute $\bfSigma_{SX} = \bfS\bfX, \bfSigma_{XX} = \bfX^\top\bfX$ and $\bfSigma_{SS}=\bfS^\top \bfS$.
}
{Compute the top generalised eigenvector of the problem
\begin{gather}
\small
\begin{bmatrix}
0 & \bfSigma_{XS}\\
\bfSigma_{XS}^\top & 0
\end{bmatrix}
\begin{bmatrix}
\bfalpha \\
\bfbeta
\end{bmatrix}
=
{\bf \rho}
\begin{bmatrix}
\bfSigma_{XX} & 0 \\
0 & \bfSigma_{SS}
\end{bmatrix}
\begin{bmatrix}
\bfalpha \\
\bfbeta
\end{bmatrix}
\end{gather}
}
\If{kernelise}
{
{Return the embeddings $\bfz_S = \bfG\bfH\bfbeta$ and $\bfz_X = \bfK\bfH\bfalpha$ that minimises upsets with respect to $\bfC$ and skill function $r(\bfx) = k(\bfx, \bfX)\bfH\bfalpha$} \;
}
\Else
{
{Return the embeddings $\bfz_S = \bfS\bfbeta$ and $\bfz_X = \bfX\bfalpha$ that minimises upsets with respect to $\bfC$ and skill function $r(\bfx) = \bfx^\top\bfalpha$. }\;
}
\caption{\textsc{CCRank}}
\label{algo: CCARank}
\end{algorithm}
\DecMargin{1em}

\section{Experiments}
\label{section: experiments}

\def\svdn{$\textsc{SVDN}$}
\def\svdc{$\textsc{SVDC}$}
\def\svdk{$\textsc{SVDK}$}
\def\serial{$\textsc{Serial}$}
\def\bt{$\textsc{BT}$}
\def\btlr{$\textsc{BT-LR}$}
\def\btgp{$\textsc{BT-GP}$}
\def\CCrank{$\textsc{CC}$}
\def\rc{$\textsc{RC}$}
\def\rrc{$\textsc{RRC}$}
\def\KCC{$\textsc{KCC}$}
\def\cserial{$\textsc{C-Serial}$}

In this section, we compare the performance of $\textsc{C-SerialRank}$ (\cserial{} from now on for brevity), \textsc{SVDCovRank} (\svdc), kernelised \textsc{SVDCovRank} (\svdk), \textsc{CCRank} (\CCrank), kernelised \textsc{CCRank} (\KCC) with that of a number of other benchmark algorithms from the literature. We compare against the spectral ranking algorithms \textsc{SerialRank} (\serial), \textsc{SVDRank} (\svdn), \textsc{Rank-Centrality} (\rc{}), \textsc{Regularised Rank-Centrality} (\rrc{}) and against the probabilistic ranking algorithms \textsc{Bradley-Terry-Luce} model (\bt) as well as its two extension based on logistic regression $(\btlr)$ and Gaussian processes \cite{chu2005preference} $(\btgp)$. For \btgp{} we will simply utilise its posterior mean for rank prediction. We note that Bradley-Terry-Luce and Rank Centrality based methods cannot handle \textit{cardinal} comparisons, therefore we first transform each entry of the pairwise comparison matrix to $\operatorname{sign}(C_{i, j})$. See Table~\ref{table: summary} for a summary of our methods and benchmarks. All code and implementations are made publicly available.\footnote{ \url{https://anonymous.4open.science/r/SpectralRankingWithCovariates-6F27}}

\begin{table}[t]
\centering
\caption{A summary of our proposed methods (shaded grey) and benchmarks.}
\resizebox{\columnwidth}{!}{
\begin{tabular}{|r||c|c|c|c|c|} \hline
\multicolumn{1}{|c||}{Algorithm} &
  \multicolumn{1}{c|}{Model type} &
  \multicolumn{1}{c|}{\begin{tabular}[c]{@{}c@{}} Comparisons\end{tabular}} &
  \multicolumn{1}{c|}{Covariates} &
  \multicolumn{1}{c|}{\begin{tabular}[c]{@{}c@{}} Unseen player \\ rank prediction \end{tabular}} \\ \hline
\textsc{BT}~\cite{bradley1952rank}                         & Probabilistic &  Ordinal & \xmark  & \xmark  \\ 
\textsc{BT-LR} $\textbackslash$ \textsc{BT-GP}~\cite{chu2005preference} & Probabilistic & Ordinal & \cmark & \cmark \\ 
\textsc{RC}~\cite{RankCentrality}                         & Spectral      & Ordinal & \xmark  & \xmark  \\ 
\textsc{RRC}~\cite{jain2020spectral}                        & Spectral       & Ordinal & \cmark & \xmark  \\ 
\textsc{Serial}~\cite{serialRank}                     & Spectral       & Ordinal + Cardinal & \xmark  & \xmark  \\ 
\rowcolor[HTML]{EFEFEF} 
\textsc{C-Serial} (Ours)                   & Spectral       & Ordinal + Cardinal & \cmark & \xmark  \\ 
\textsc{SVD}~\cite{cucuringu2016simple}                        & Spectral       & Ordinal + Cardinal & \xmark  & \xmark  \\ 
\rowcolor[HTML]{EFEFEF} 
\textsc{SVDC} $\textbackslash$ \textsc{SVDK} (Ours)  & Spectral       & Ordinal + Cardinal & \cmark & \cmark \\ 
\rowcolor[HTML]{EFEFEF} 
\textsc{CC} $\textbackslash$ \textsc{KCC} (Ours)     & Spectral       & Ordinal + Cardinal & \cmark & \cmark \\ \hline
\end{tabular}
}
\label{table: summary}
\end{table}

Throughout our experiments, we will use the Radial Basis Function (RBF) kernel $k(x,x') = \exp\left(\frac{||x-x'||^2}{2\ell_k}\right)$ where the lengthscale $\ell_k$ is chosen via the median heuristic method. Hyperparameters such as $\lambda$ in \cserial{} are tuned using 10-fold cross-validation on withheld match outcomes. In practice, we should select the kernel that reflects the nature of the non-linearity in data. We assess the performance of all algorithms via the following two tasks:

\paragraph{Task 1: Rank Inference.} Given $n$ players with a sparse and potentially noisy pairwise comparison matrix $\bfC$, we pick $\textbf{70\%}$ \textbf{of the observed matches} as training data $\bfC^{(1)}_{\text{train}}$, and the remaining observed matches as testing data $\bfC^{(1)}_{\text{test}}$. Our goal is to train the algorithms on $\bfC^{(1)}_{\text{train}}$ to recover the rankings, and then measure their performances by reporting the accuracy scores in predicting matches from $\bfC^{(1)}_{\text{test}}$ based on the estimated rankings.

\paragraph{Task 2: Rank Prediction.} Given $n$ players with comparison matrix $\bfC$, withhold $\mathbf{70\%}$ \textbf{of the players} and observed the induced comparison matrix $\bfC^{(2)}_{\text{train}}\in \RR^{n' \times n'}$ where $n' = \lfloor n \times 0.7 \rfloor$. Set up the testing data $\bfC^{(2)}_{\text{test}}$ based on the induced comparison matrix of the remaining players. We train on $\bfC^{(2)}_{\text{train}}$ and report the accuracy in predicting matches from $\bfC^{(2)}_{\text{test}}$ based on the predicted rankings on unseen players. 


\subsection{Simulations on synthetic data}

We test the performances of the 12 ranking algorithms on both rank inference and prediction problems using synthetic data. We report the Kendall Tau score on test data as our performance metric, as we have access to the ground truth ranking. We simulate $n=1000$ players with covariates $\bfX$ drawn from $N(0, 1)$. We set the skill function as $r(x) = \sin(3\pi x) - 1.5x^2 + \epsilon N(0, 1)$ with the noise level $\epsilon \in \{0, 0.05, 0.2\}$. The corresponding comparison matrix $\bfC$ is given by $C_{i, j} = r(x_i) - r(x_j)$. Besides feature noise, we also consider adding noises to the matches. We follow the \textit{FLIP} noise model deployed in~\cite{syncRank} where we flip each match result with probability $p \in \{0, 0.05, 0.1, 0.2, 0.3, 0.4, 0.5\}$. With probability $\phi \in \{0.7, 0.9\}$, we remove an observed match in the comparison matrix $\bfC$ to yield sparse comparisons matrices as in most practical situations. Experiments are repeated over 10 seeds.

\begin{figure}
    \centering
    \subfloat[][Rank Inference]{
    \includegraphics[width=0.98\columnwidth]{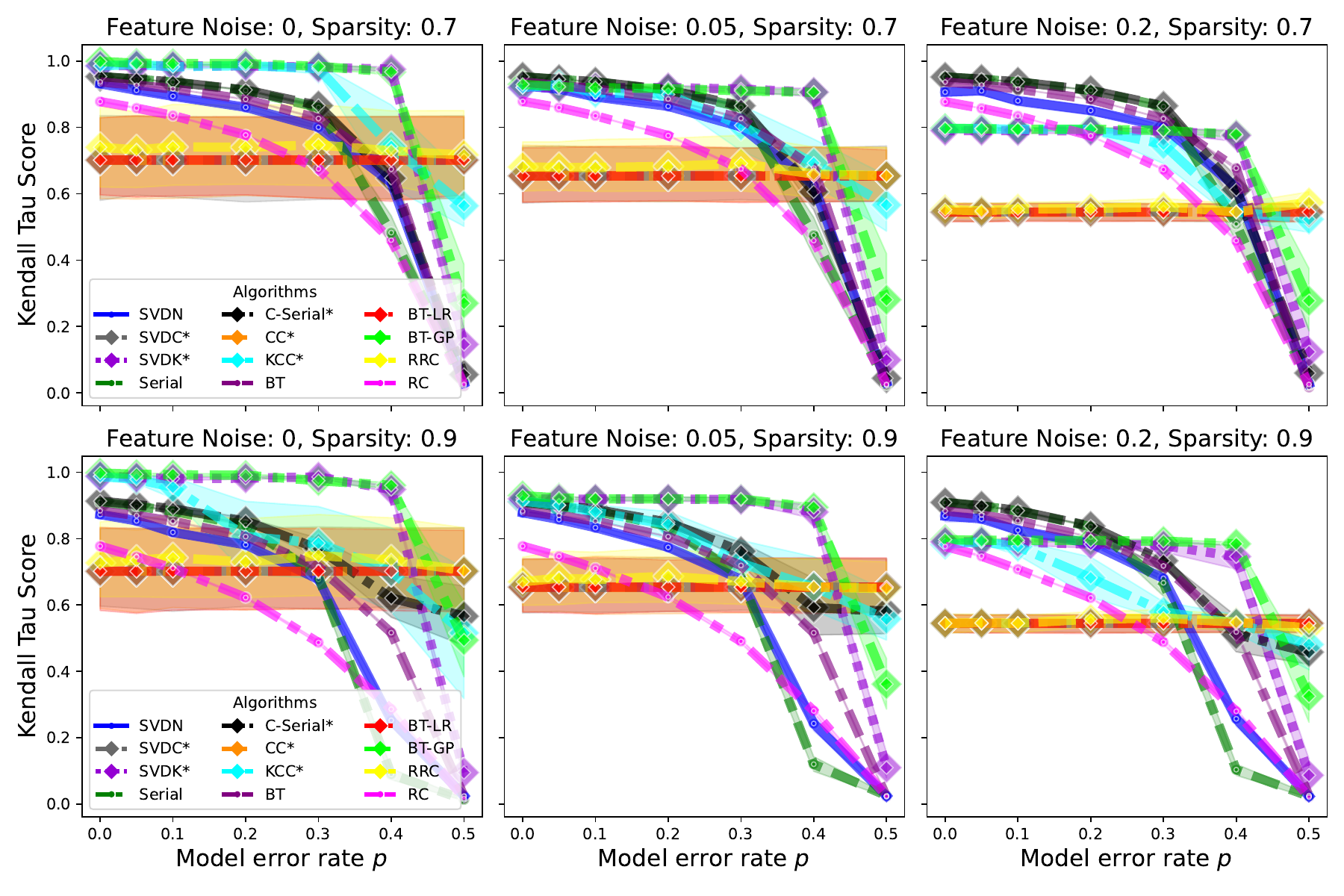}
    \label{fig: simulations-rank-inference}
    } \\
    \subfloat[][Rank Prediction]{
    \includegraphics[width=0.98\columnwidth]{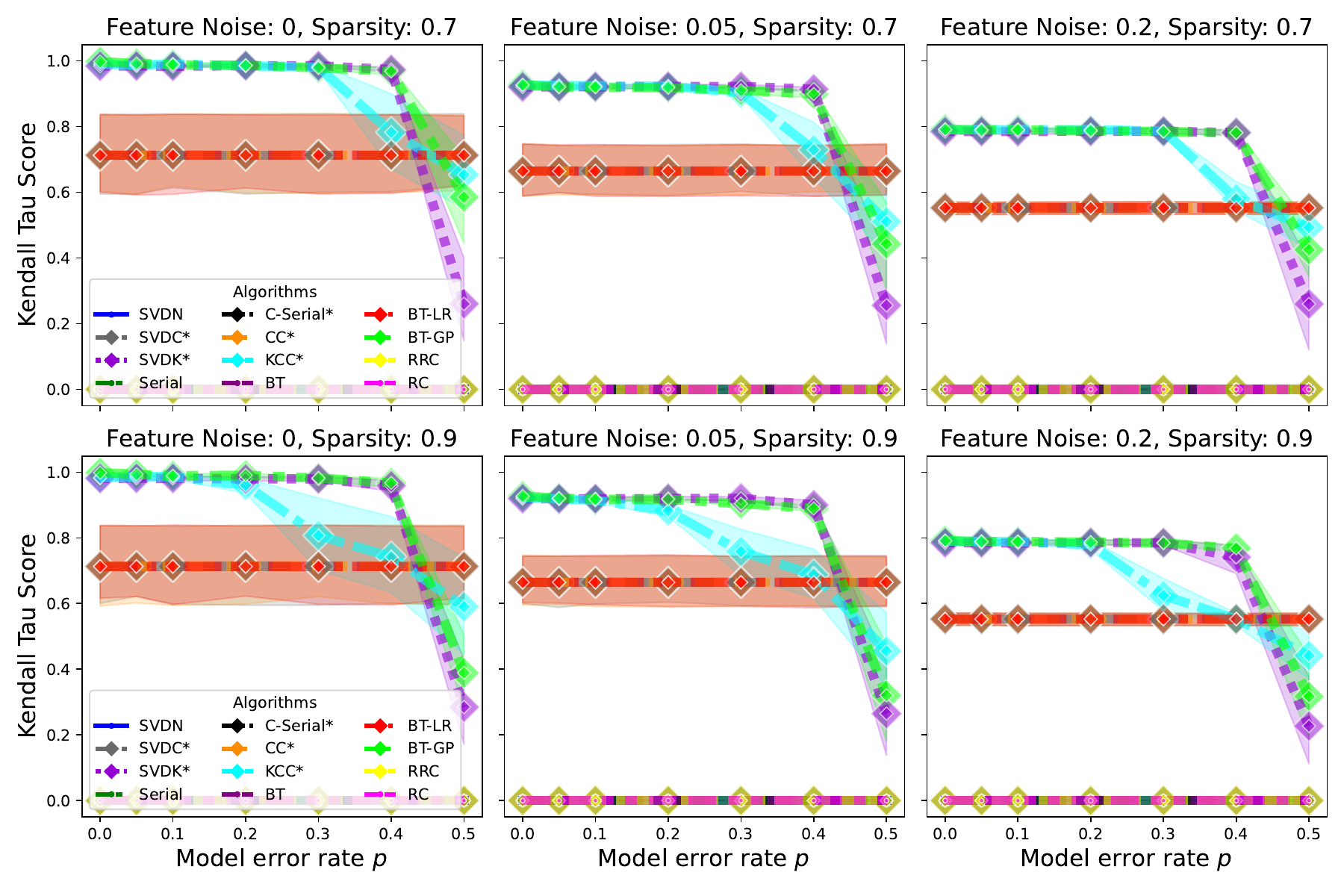}
    \label{fig: simulations-rank-prediction}
    }    
    \caption{Performance in terms of Kendall's tau score (the closer to $1$, the better) under different synthetic noise models with $1000$ players. Results are averaged over $10$ iterations and $95\%$ confidence intervals are reported. Methods with diamond markers use covariates, and $*$ indicates our proposed methods.}
    \label{fig:my_label}
\end{figure}

Fig.~\ref{fig: simulations-rank-inference} and Fig.~\ref{fig: simulations-rank-prediction} demonstrate the performances of the algorithms under different sparsity levels, feature noises and model error rates for the \textit{inference} and \textit{prediction} tasks. For the \textit{inference} tasks, we see a performance drop in the covariate-based ranking algorithms as the feature noise increases. This coincides with our usual ``garbage covariate-in, garbage out''  intuition in modelling. We also note that as the skill vector non-linearly relates to the covariates by design, all kernelised algorithms outperformed their linear counterparts. We also observe that as the flip error rate $p$ increases, the performance of covariate-free algorithms decreases quickly, while covariate-based methods performed relatively stable until the extreme error rate. For \textit{prediction} tasks, similar trends to the \textit{inference} tasks in terms of algorithmic performances can be seen with respect to model error rates, sparsity and feature noise levels. 


Specifically, we observe that the kernelised spectral ranking algorithm \svdk{} performed competitively to probabilistic ranking \btgp{} until a high flip error rate in both tasks. \KCC{} performed slightly worse than the two, but still much better than the covariate-based spectral method baseline \rrc{} in general. We note that the linear methods do not appear to be sensitive to model error rate because the ranking extracted is dominated by the information in the covariates.


%




\begin{table}[t]
\centering
\caption{Accuracy of match outcome predictions based on recovered ranking on each data set. Scores are averaged over 10 seeds, and 1 s.d is reported. $\dagger$ indicates algorithm using covariate information. We shaded our proposed methods in grey. The best performing method is shown in bold red, and second best in bold blue.}
\resizebox{\columnwidth}{!}{
{\small
\begin{tabular}{|r||cc|cc|cc|cc|} \hline
\multicolumn{1}{|r||}{} &
  \multicolumn{2}{c|}{\underline{Chameleon}} &
  \multicolumn{2}{c|}{\underline{FlatLizard}} &
  \multicolumn{2}{c|}{\underline{Pokemon}} &
  \multicolumn{2}{c|}{\underline{NFL 2000-18}}  \\
\multicolumn{1}{|r||}{Algorithm} &
  \multicolumn{1}{c}{Task 1} &
  \multicolumn{1}{c|}{Task 2} &
  \multicolumn{1}{c}{Task 1} &
  \multicolumn{1}{c|}{Task 2} &
  \multicolumn{1}{c}{Task 1} &
  \multicolumn{1}{c|}{Task 2} &
  \multicolumn{1}{c}{Task 1} &
  \multicolumn{1}{c|}{Task 2}\\ \hline 
\textsc{BT} &
  \multicolumn{1}{c|}{\textbfr{$\mathbf{0.83_{\pm 0.03}}$}} &
  \multicolumn{1}{c|}{-} &
  \multicolumn{1}{c|}{\textbfb{$0.86_{\pm0.06}$}} &
  \multicolumn{1}{c|}{-} &
  \multicolumn{1}{c|}{$0.86_{\pm0.01}$} &
  \multicolumn{1}{c|}{-} &
  \multicolumn{1}{c|}{$0.63_{\pm0.04}$} &
  \multicolumn{1}{c|}{-}  \\
\textsc{BT-LR}$^\dagger$ &
  \multicolumn{1}{c|}{$0.71_{\pm0.03}$} &
  \textbfb{$0.82_{\pm0.14}$} &
  \multicolumn{1}{c|}{$0.84_{\pm0.04}$} &
  \textbfb{$0.77_{\pm0.12}$} &
  \multicolumn{1}{c|}{\textbfb{$0.88_{\pm0.01}$}} &
  \multicolumn{1}{c|}{\textbfb{$0.89_{\pm0.01}$}} &  \multicolumn{1}{c|}{\textbfr{$\mathbf{0.68_{\pm0.04}}$}} & 
  \textbfb{$0.62_{\pm0.09}$} \\
\textsc{BT-GP}$^\dagger$ &
  \multicolumn{1}{c|}{$0.75_{\pm0.04}$} &
  $0.74_{\pm0.13}$ &
  \multicolumn{1}{c|}{$0.80_{\pm0.05}$} &
$0.74_{\pm0.12}$  &
  \multicolumn{1}{c|}{$0.72_{\pm0.01}$} &
$0.72_{\pm0.02}$ &
  \multicolumn{1}{c|}{$0.58_{\pm0.04}$} &
$0.61_{\pm0.08}$ \\
\textsc{RC} &
  \multicolumn{1}{c|}{$0.61_{\pm 0.06}$} &
  \multicolumn{1}{c|}{-} &
  \multicolumn{1}{c|}{$0.66_{\pm0.05}$} &
  \multicolumn{1}{c|}{-} &
  \multicolumn{1}{c|}{$0.81_{\pm0.01}$} &
  \multicolumn{1}{c|}{-} &
  \multicolumn{1}{c|}{$0.58_{\pm0.04}$} &
  \multicolumn{1}{c|}{-}  \\
  \textsc{RRC}$^\dagger$ &
  \multicolumn{1}{c|}{$0.61_{\pm 0.03}$} &
  \multicolumn{1}{c|}{-} &
  \multicolumn{1}{c|}{$0.66_{\pm0.01}$} &
  \multicolumn{1}{c|}{-} &
  \multicolumn{1}{c|}{$0.85_{\pm0.01}$} &
  \multicolumn{1}{c|}{-} &
  \multicolumn{1}{c|}{$0.62_{\pm0.04}$} &
  \multicolumn{1}{c|}{-}  \\
\textsc{SVD} &
  \multicolumn{1}{c|}{$0.72_{\pm0.08}$} &
  \multicolumn{1}{c|}{-} &
  \multicolumn{1}{c|}{$0.69_{\pm0.05}$} &
  \multicolumn{1}{c|}{-} &
 \multicolumn{1}{c|}{$0.84_{\pm0.02}$} &
  \multicolumn{1}{c|}{-} &
    \multicolumn{1}{c|}{$0.57_{\pm0.04}$} &
 \multicolumn{1}{c|}{-}\\
 \rowcolor[HTML]{EFEFEF} 
\textsc{SVDC}$^\dagger$ &
  \multicolumn{1}{c|}{$0.65_{\pm0.06}$} &
$0.74_{\pm0.12}$ &
  \multicolumn{1}{c|}{$0.81_{\pm0.05}$} &
$0.73_{\pm0.10}$ &
  \multicolumn{1}{c|}{\textbfr{$\mathbf{0.89_{\pm0.01}}$}} &
  \multicolumn{1}{c|}{\textbfr{$\mathbf{0.90_{\pm0.01}}$}} &
  \multicolumn{1}{c|}{$0.66_{\pm0.04}$} &
\textbfb{$0.62_{\pm0.09}$}  \\
\rowcolor[HTML]{EFEFEF} %
\textsc{SVDK}$^\dagger$ &
  \multicolumn{1}{c|}{$0.76_{\pm0.06}$} &
  $0.79_{\pm0.06}$ &
  \multicolumn{1}{c|}{$0.68_{\pm0.05}$} &
  $0.65_{\pm0.09}$&
  \multicolumn{1}{c|}{$0.85_{\pm0.01}$} &
  \multicolumn{1}{c|}{$0.72_{\pm0.09}$} &
    \multicolumn{1}{c|}{$0.59_{\pm0.04}$} &
\textbfb{$0.62_{\pm0.09}$} \\
\textsc{Serial} &
  \multicolumn{1}{c|}{$0.79_{\pm0.04}$} &
  \multicolumn{1}{c|}{-} &
  \multicolumn{1}{c|}{$0.70_{\pm0.05}$} &
  \multicolumn{1}{c|}{-} &
  \multicolumn{1}{c|}{\textbfb{$0.88_{\pm0.01}$}} &
  \multicolumn{1}{c|}{-} &
    \multicolumn{1}{c|}{$0.58_{\pm0.04}$} &
\multicolumn{1}{c|}{-}\\
\rowcolor[HTML]{EFEFEF} 
\textsc{C-Serial}$^\dagger$ &
  \multicolumn{1}{c|}{\textbfb{$0.80_{\pm0.03}$}} &
  \multicolumn{1}{c|}{-} &
  \multicolumn{1}{c|}{\textbfr{$\mathbf{0.88_{\pm0.01}}$}} &
  \multicolumn{1}{c|}{-} &
  \multicolumn{1}{c|}{\textbfb{$0.88_{\pm0.01}$}} &
  \multicolumn{1}{c|}{-} & 
\multicolumn{1}{c|}{$0.59_{\pm0.04}$} &
  \multicolumn{1}{c|}{-}  \\ 
\rowcolor[HTML]{EFEFEF} 
\textsc{CC}$^\dagger$ &
  \multicolumn{1}{c|}{$0.66_{\pm0.10}$} &
  \textbfr{$\mathbf{0.95_{\pm0.10}}$} &
  \multicolumn{1}{c|}{$0.78_{\pm0.08}$} &
  \textbfr{$\mathbf{0.92_{\pm0.15}}$} &
  \multicolumn{1}{c|}{$0.78_{\pm0.08}$} &
  \multicolumn{1}{c|}{$0.79_{\pm0.08}$} &
  \multicolumn{1}{c|}{$0.61_{\pm0.05}$} &
  \multicolumn{1}{c|}{\textbfr{$\mathbf{0.66_{\pm0.10}}$}}\\
\rowcolor[HTML]{EFEFEF} 
\textsc{KCC}$^\dagger$ &
  \multicolumn{1}{c|}{$0.71_{\pm0.06}$} &
  $0.80_{\pm0.10}$ &
  \multicolumn{1}{c|}{$0.78_{\pm0.03}$} &
  $0.71_{\pm0.11}$ &
  \multicolumn{1}{c|}{$0.81_{\pm0.01}$} &
  \multicolumn{1}{c|}{$0.72_{\pm0.09}$} &
  \multicolumn{1}{c|}{\textbfb{$0.67_{\pm0.05}$}}&
  \multicolumn{1}{c|}{\textbfb{$0.62_{\pm0.08}$}} \\ \hline
\end{tabular}
\label{table: real-world}
}}
\end{table}

\subsection{Experiments on real-world data}

\begin{figure}

    \centering
    \subfloat[][Chameleon$\ddag$]{
    \includegraphics[width=0.5\textwidth]{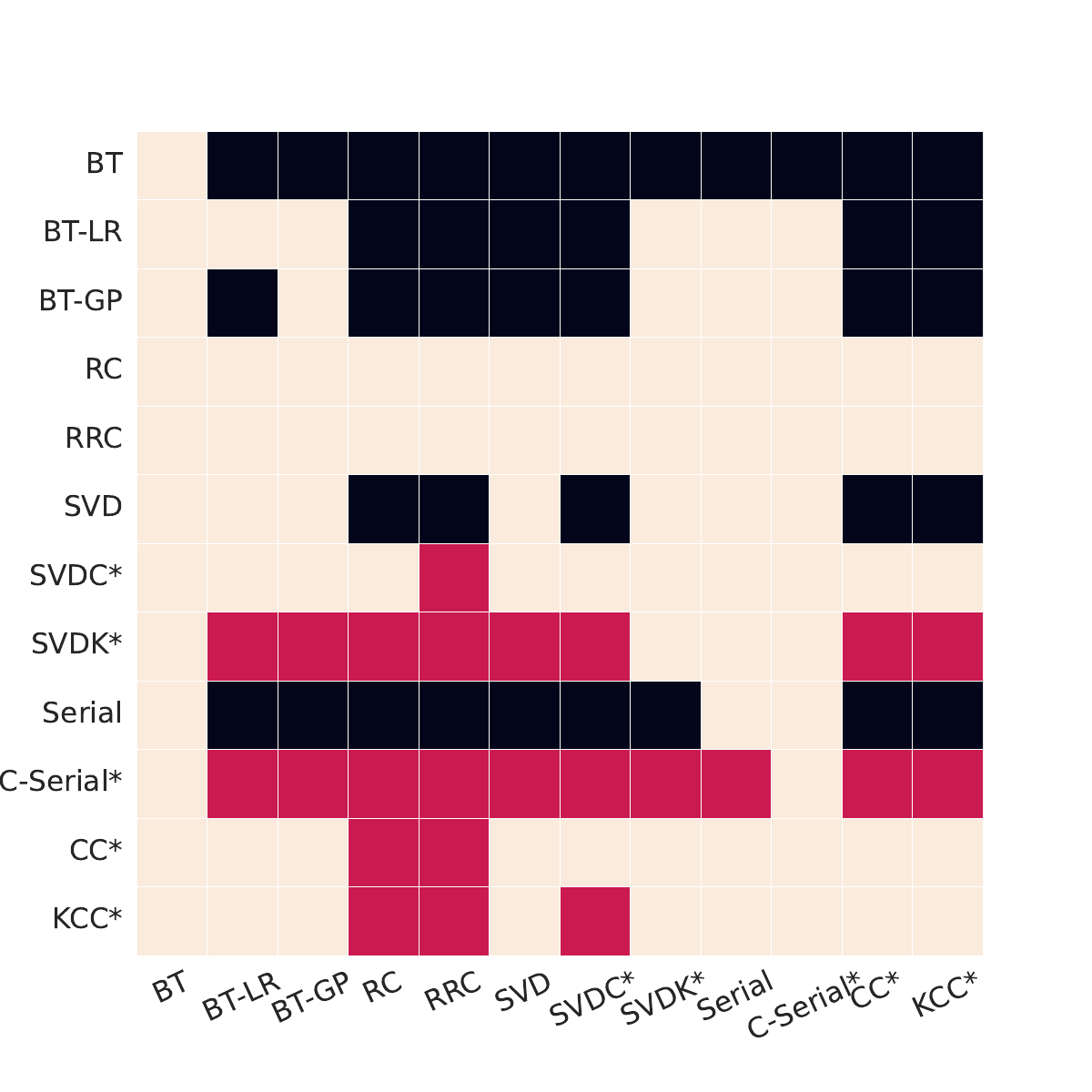}
    }
    \subfloat[][Flatlizard$\ddag$]{
    \includegraphics[width=0.5\textwidth]{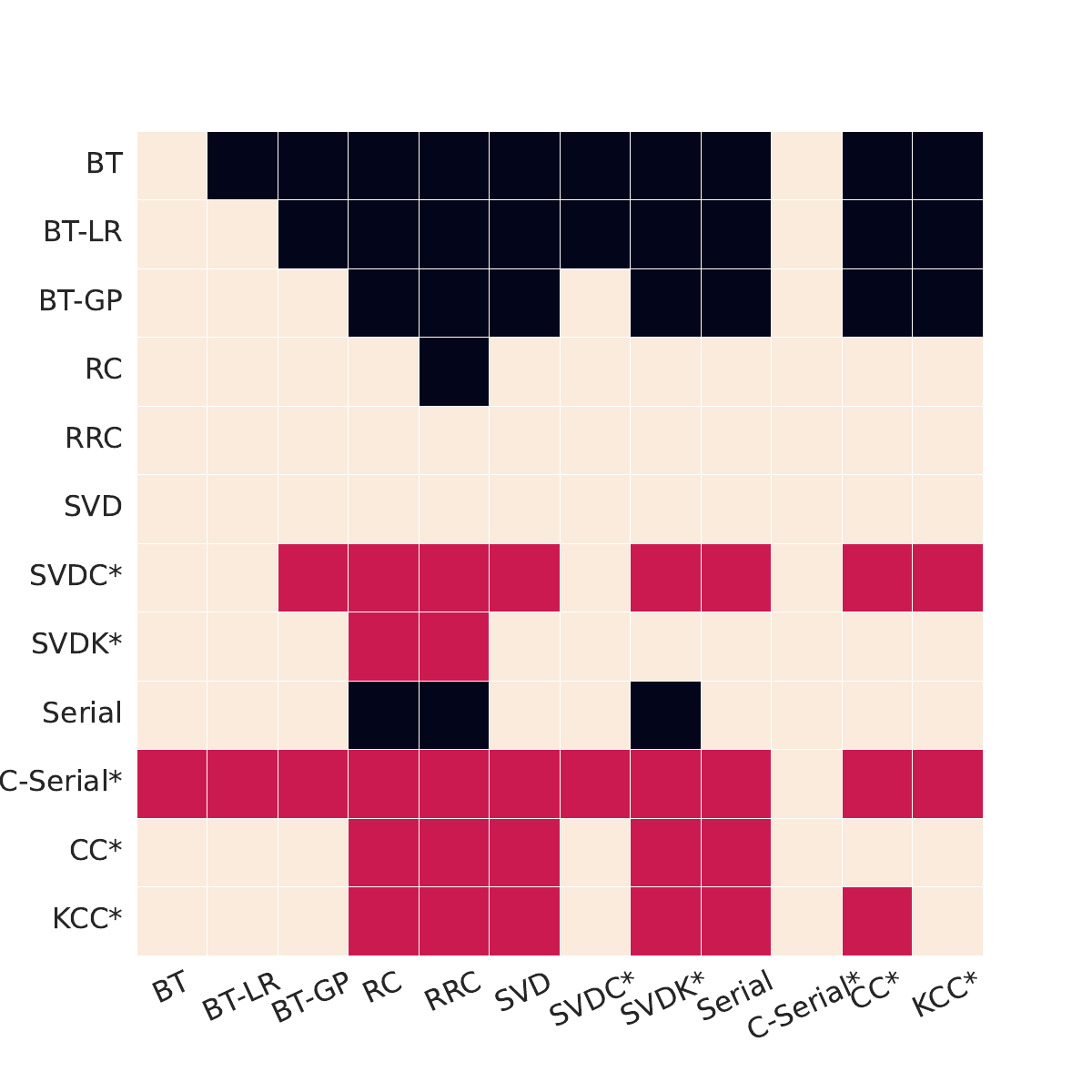}
    }\\
    \subfloat[][Pokemon$\ddag$]{
    \includegraphics[width=0.5\textwidth]{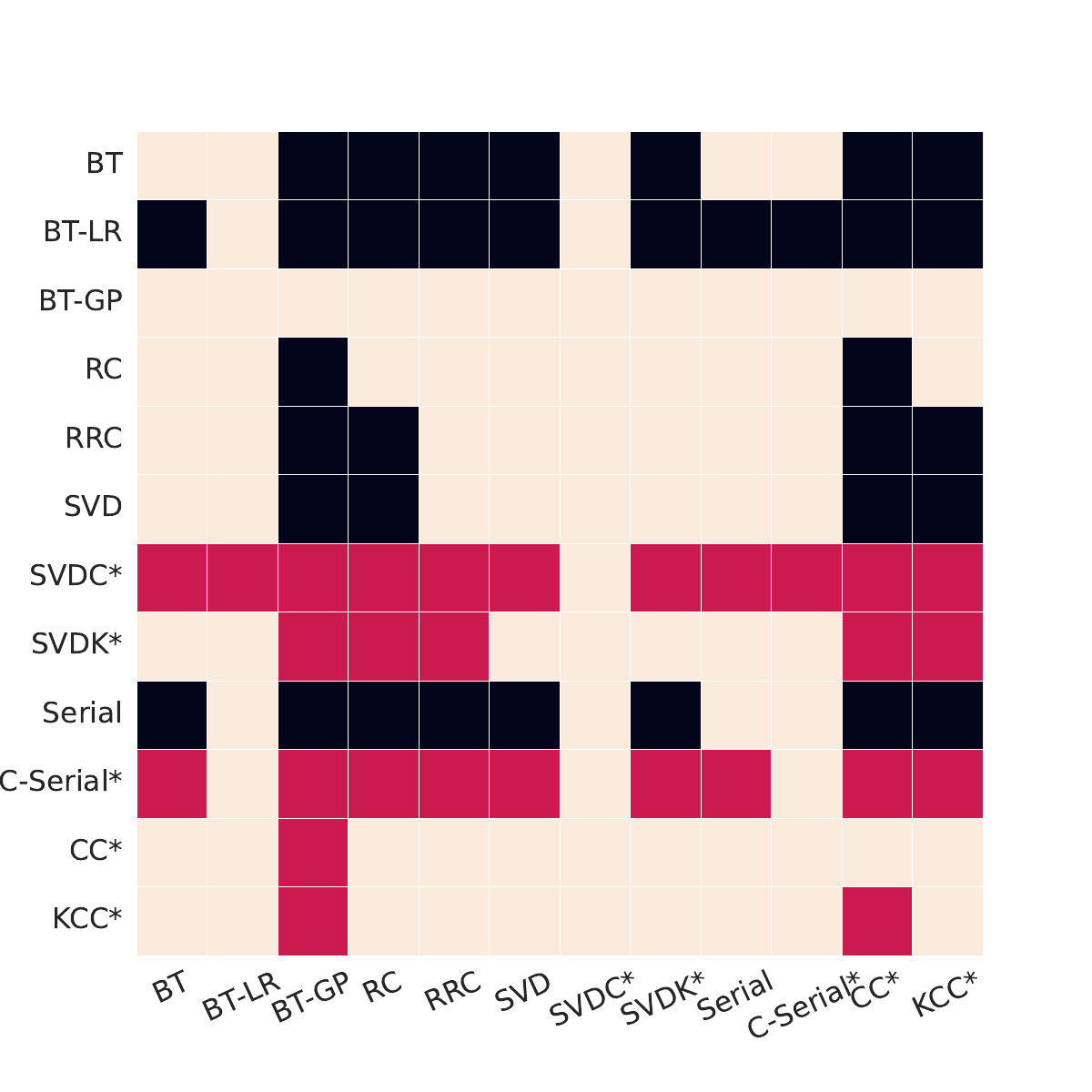}
    }
    \subfloat[][NFL $\ddag$]{
    \includegraphics[width=0.5\textwidth]{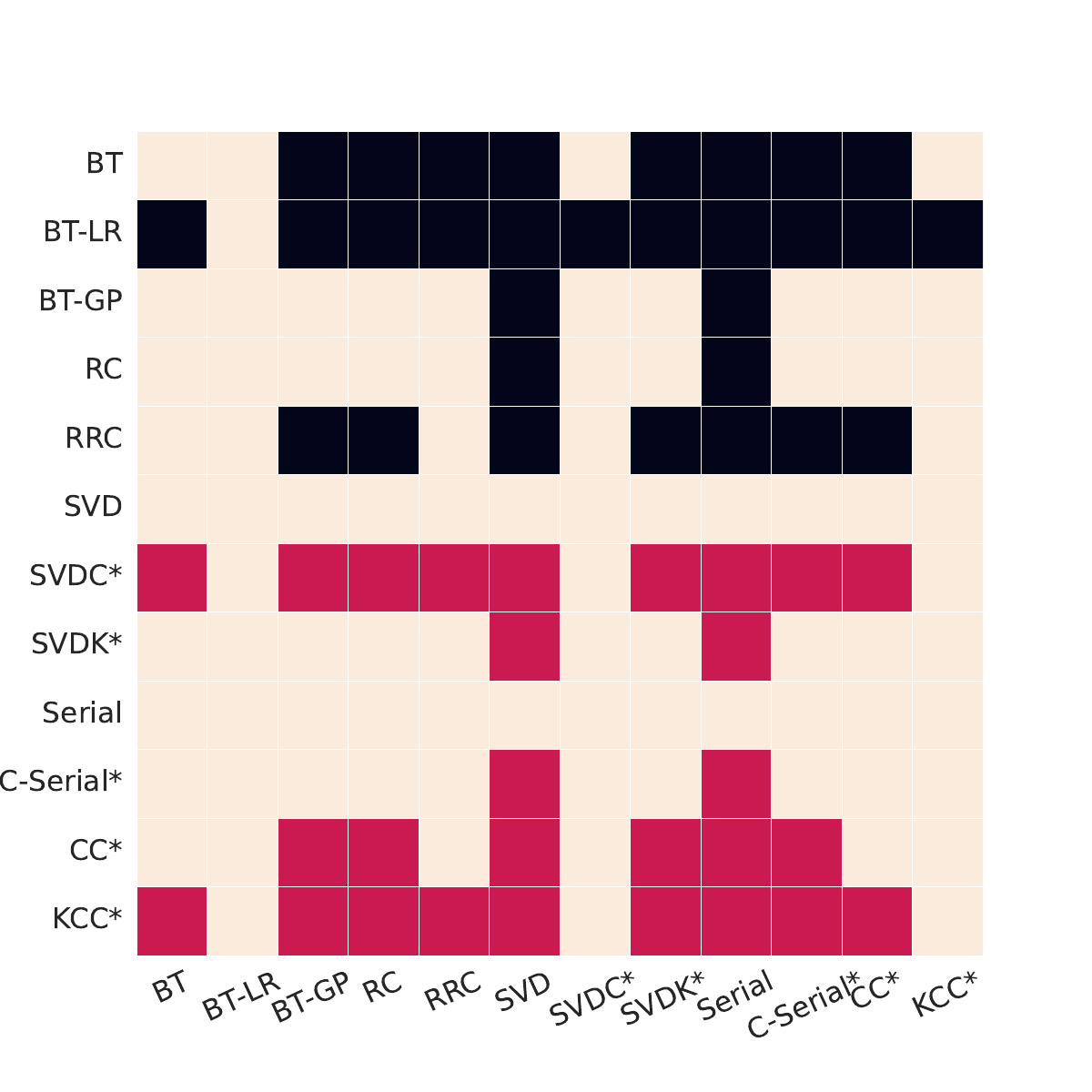}
    }\\
    \resizebox{\columnwidth}{!}{
    \subfloat[][Chameleon$\mathparagraph$]{
    \includegraphics[width=0.25\textwidth]{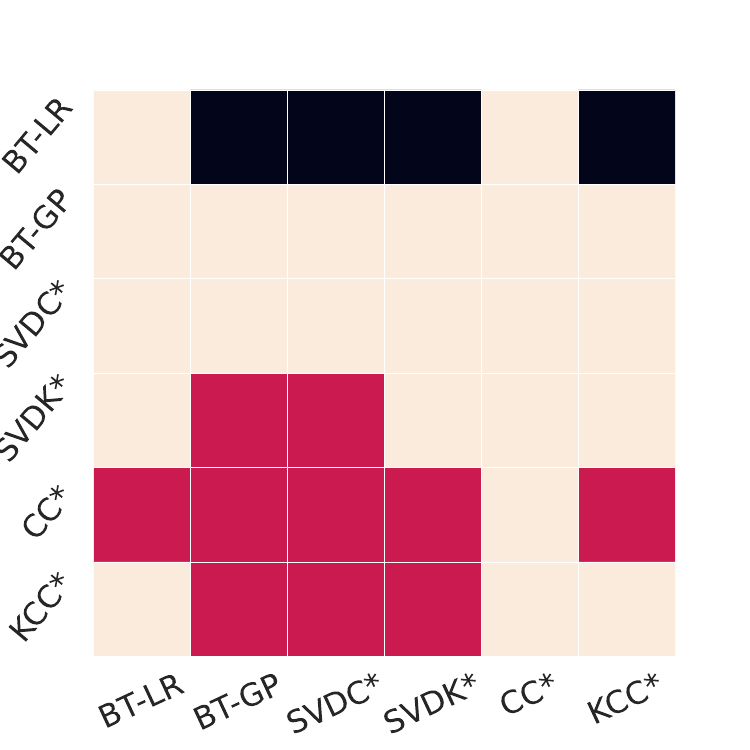}
    }
    \subfloat[][Flatlizard$\mathparagraph$]{
    \includegraphics[width=0.25\textwidth]{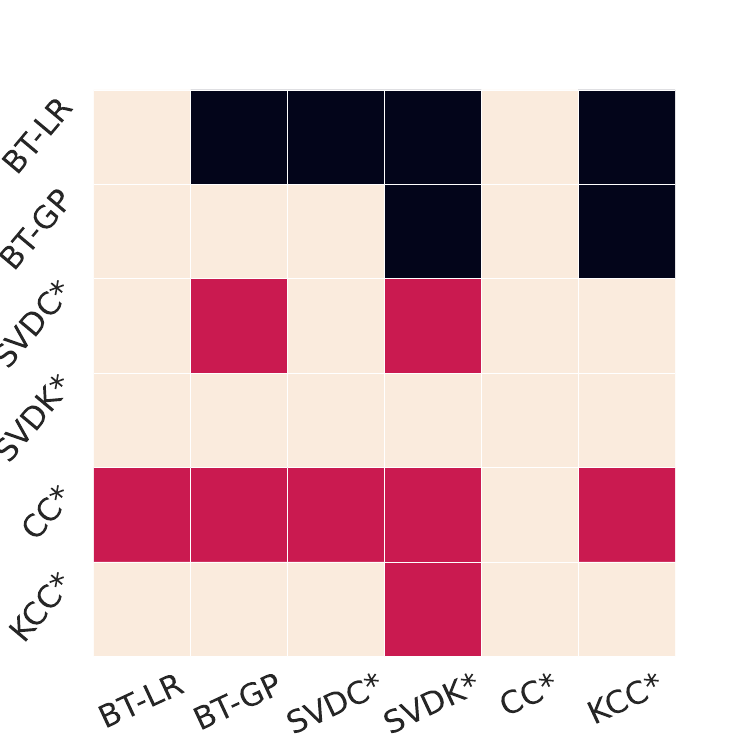}
    }
    \subfloat[][Pokemon$\mathparagraph$]{
    \includegraphics[width=0.25\textwidth]{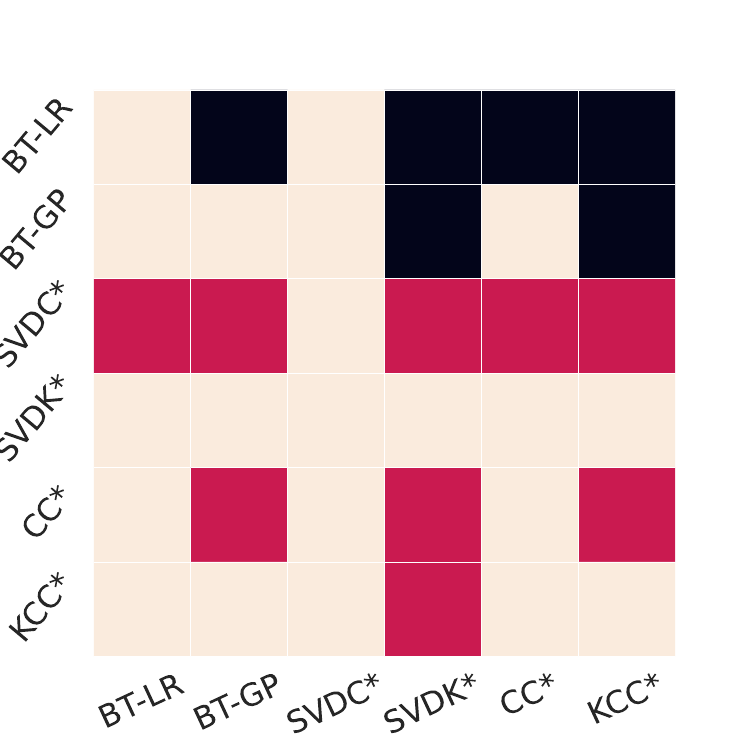}
    }
    \subfloat[][NFL$\mathparagraph$]{
    \includegraphics[width=0.25\textwidth]{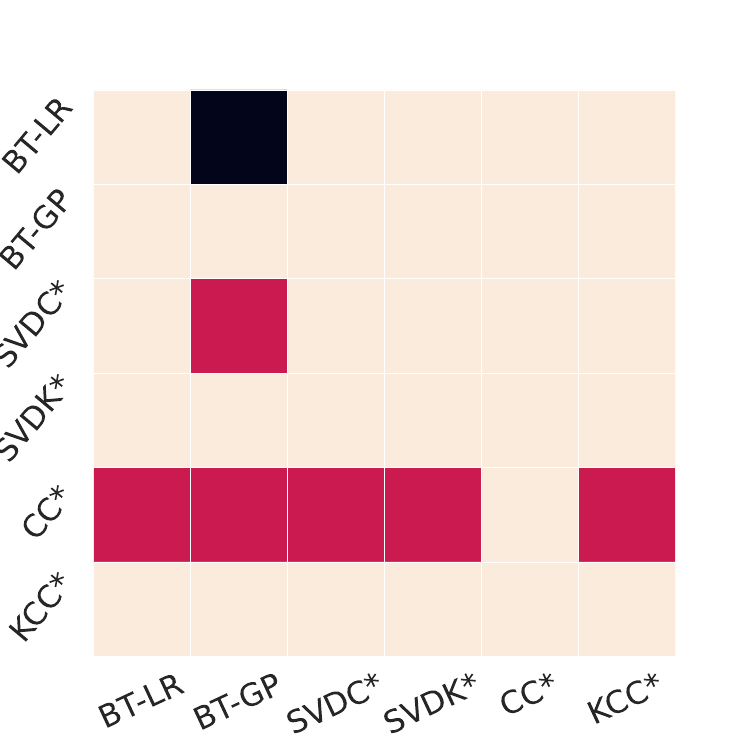}
    }
    }
    \caption{Statistical Significance plots for our real-world experiments. We shade an entry if the $i^{th}$ row algorithm is statistically significantly better than the $j^{th}$ column algorithm on a one-sided paired Wilcoxon test with a 5\% significance level. We shade our algorithms in red and the benchmarks in black. $\ddag$ and $\mathparagraph$ indicate rank inference and prediction tasks respectively.}
    \label{fig: sigleveltask1}
\end{figure}

In this section, we apply all ranking algorithms to a variety of real-world data sets ($22$ in total) to perform rank \textit{inferences} and \textit{predictions}. We average the accuracy scores over 10 seeds and report the results in Table~\ref{table: real-world}. $5\%$ significant level one-sided paired Wilcoxon tests for all algorithm pairs are conducted for each data set, with results reported in Fig.~\ref{fig: sigleveltask1}.


\paragraph{Male Cape Dwarf Chameleons Contest.} The first data set is used in the study by Stuart-Fox et al.~\cite{stuart2006multiple}. Physical measurements are made on $35$ male Cape dwarf chameleons, and the results of $104$ contests are measured and encoded as ordinal comparison data. From Table~\ref{table: real-world} and Fig.~\ref{fig: sigleveltask1} we see that \bt{} outperformed the rest in rank \textit{inferences} (closely followed by $\cserial{}$), while \CCrank{} performed best compared with other covariate based methods in rank \textit{prediction}.

\paragraph{Flatlizard Competition.} The data is collected at Augrabies Falls National Park (South Africa) in September-October $2002$ \cite{WM09}, on the contest performance and background attributes of $77$ male flat lizards (\emph{Platysaurus Broadleyi}). The results of $100$ contests were recorded, along with $18$ physical measurements made on each lizard, such as \textit{weight} and \textit{head size}. From Fig~\ref{fig: sigleveltask1} we see that \cserial{}  and \CCrank{} statistically significantly outperformed the rest for inference and prediction, respectively.

\paragraph{Pokémon Battle.} Our next data set comes from a Kaggle competition\footnote{data collected from \url{https://www.kaggle.com/c/intelygenz-pokemon-challenge}}. Approximately $45,000$ battles among $800$ Pokémon are recorded in ordinal format. $25$ characteristic of each Pokémon, such as their \textit{type}, \textit{attack} and \textit{defence}, are available. These battles are generated by a custom algorithm written by the Kaggle host that closely reflects the game mechanics. Table~\ref{table: real-world} shows that linear covariate-based models, in general, outperformed their kernelised counterparts. This is not surprising as the ``strength'' of a pokemon is linearly related to the covariates provided (e.g. \textit{attack} and \textit{defence)}. Given these highly informative features, covariate-based methods, in general, outperformed covariate-free methods as well. We note that \svdc{}, closely followed by \btlr{}, was the top performer for both inference and prediction tasks.

\paragraph{NFL Football 2000-2018.} The last data set we use contains the outcome of National Football League (NFL) matches during the regular season, for the years $2000-2018$\footnote{data collected from \url{nfl.com}}. In addition, $256$ matches per year between $32$ teams, along with $18$ performance metrics, such as \textit{yards per game} and \textit{number of fumbles} are recorded. We used their score differences at each match as the cardinal comparisons. We run our algorithms on each year's comparison graphs separately and average the results. We see for rank \textit{inference}, our covariate-based algorithms outperformed their covariate-free counterparts, with \btlr{} the top performer, followed closely by \svdc{}, \KCC{} and \rrc{}. For rank \textit{prediction}, \CCrank{} was significantly better than all other methods.



\vspace{-0.3cm}
\section{Conclusions and Future directions}
\label{section: discussion}

We proposed three spectral ranking algorithms to tackle the problem of ranking given noisy and incomplete pairwise comparisons when player covariates are available. We  demonstrated the efficacy, strengths and weaknesses of the proposed methods in comparison to baselines through an extensive set of numerical experiments on synthetic and $22$ real-world problems. 

\textsc{C-SerialRank} extends \textsc{SerialRank} by incorporating covariate-induced player similarities into the seriation process through a linear combination. Although simple, through a straight forward cross-validation scheme on the regularisation parameter, we can ensure the ranking recovered in this way is at least as good as the ranking recovered from \textsc{SerialRank}, as demonstrated in our experiments.  The downside of \textsc{C-SerialRank}, similar to \textsc{Regularised Rank Centrality}, is that it cannot be used on rank prediction. However, the results of our experiments show that the former outperformed the latter in the majority of the experiments.  Our second contribution \textsc{SVDCovRank} extends \textsc{SVDRank} by relating covariates to the eigenvector of the pairwise comparison matrix $\bfC$ in a regression fashion. This allows us to predict the rankings of unseen players, which previous spectral ranking with covariates cannot handle. As showed in our experiments, \textsc{SVDCovRank} performed comparably to covariate-based probabilistic methods. At last, \textsc{CCRank} approaches the ranking problem using the canonical correlation analysis between match outcomes and player covariates. \textsc{CCRank} places fewer assumptions on the underlying data generating process and is thus more robust in practice. This is demonstrated by its competitive performance against other methods when predicting the rank of unseen players.

Each proposed method will find its application in different practical scenarios. When rank prediction on unseen players is not a concern, we recommend practitioners to use the \textsc{C-SerialRank} algorithm because of its simplicity in balancing covariate and match information. On the other hand, if one believes the pairwise comparisons are caused by a strong underlying skill difference, then \textsc{SVDCovRank} algorithms should be considered. At last, if one seeks to place fewer assumptions on the data, then the \textsc{CCRank} algorithm may give a more robust ranking than the approaches that place specific structures on the match outcome matrix $\bfC$. 


There are several avenues for future work. An interesting direction would be to incorporate more general types of side information into the ranking problem, e.g., prior information on partially observed player relationships. Yet another relevant direction is the extraction  of partial rankings using the side covariates. In many real-world scenarios, aiming for a global ordering of the players is unrealistic, and one is often interested in uncovering partial orderings that reflect accurately various subsets/clusters of a less homogeneous population of players. Leveraging features for the discovery of latent structures behind cyclical and inconsistent preferences, as in \cite{chau2022learning} but using spectral methods, is also an interesting direction to explore.



\bibliographystyle{splncs04}
\bibliography{main}

\begin{thebibliography}{10}
\providecommand{\url}[1]{\texttt{#1}}
\providecommand{\urlprefix}{URL }
\providecommand{\doi}[1]{https://doi.org/#1}

\bibitem{atkins1998spectral}
Atkins, J.E., Boman, E.G., Hendrickson, B.: A spectral algorithm for seriation
  and the consecutive ones problem. SIAM Journal on Computing  \textbf{28}(1),
  297--310 (1998)

\bibitem{bradley1952rank}
Bradley, R.A., Terry, M.E.: Rank analysis of incomplete block designs: I. the
  method of paired comparisons. Biometrika  \textbf{39}(3/4),  324--345 (1952)

\bibitem{caron2012efficient}
Caron, F., Doucet, A.: Efficient bayesian inference for generalized
  bradley--terry models. Journal of Computational and Graphical Statistics
  \textbf{21}(1),  174--196 (2012)

\bibitem{cattelan2012models}
Cattelan, M.: Models for paired comparison data: A review with emphasis on
  dependent data. Statistical Science  \textbf{27}(3),  412--433 (2012)

\bibitem{chau2022learning}
Chau, S., Gonz{\'a}lez, J., Sejdinovic, D.: Learning inconsistent preferences
  with gaussian processes  (2022)

\bibitem{chen2016}
Chen, S., Joachims, T.: Modeling intransitivity in matchup and comparison data.
  In: Proceedings of the Ninth ACM International Conference on Web Search and
  Data Mining. p. 227–236 (2016)

\bibitem{chu2005preference}
Chu, W., Ghahramani, Z.: Preference learning with {G}aussian processes. In:
  Proceedings of the 22nd International Conference on Machine Learning. pp.
  137--144 (2005)

\bibitem{syncRank}
Cucuringu, M.: {Sync-Rank: Robust Ranking, Constrained Ranking and Rank
  Aggregation via Eigenvector and Semidefinite Programming Synchronization}.
  IEEE Transactions on Network Science and Engineering  \textbf{3}(1),  58--79
  (2016)

\bibitem{cucuringu2016simple}
Cucuringu, M., Koutis, I., Chawla, S., Miller, G., Peng, R.: Simple and
  scalable constrained clustering: a generalized spectral method. In:
  Artificial Intelligence and Statistics. pp. 445--454 (2016)

\bibitem{SVDRank}
d'Aspremont, A., Cucuringu, M., Tyagi, H.: Ranking and synchronization from
  pairwise measurements via svd. Journal of Machine Learning Research
  \textbf{22}(19) (2021)

\bibitem{serialRank}
Fogel, F., d'Aspremont, A., Vojnovic, M.: Serialrank: Spectral ranking using
  seriation. Advances in Neural Information Processing Systems  \textbf{27}
  (2014)

\bibitem{fukumizu2007statistical}
Fukumizu, K., Bach, F.R., Gretton, A.: Statistical consistency of kernel
  canonical correlation analysis. Journal of Machine Learning Research
  \textbf{8}(Feb),  361--383 (2007)

\bibitem{hotelling1992relations}
Hotelling, H.: Relations between two sets of variates. In: Breakthroughs in
  statistics, pp. 162--190. Springer (1992)

\bibitem{huang2006generalized}
Huang, T.K., Weng, R.C., Lin, C.J., Ridgeway, G.: Generalized bradley-terry
  models and multi-class probability estimates. Journal of Machine Learning
  Research  \textbf{7}(1) (2006)

\bibitem{jain2020spectral}
Jain, L., Gilbert, A., Varma, U.: Spectral methods for ranking with scarce
  data. In: Conference on Uncertainty in Artificial Intelligence. pp. 609--618.
  PMLR (2020)

\bibitem{KendallSmith1940}
Kendall, M.G., Smith, B.B.: On the method of paired comparisons. Biometrika
  \textbf{31}(3-4),  324--345 (1940)

\bibitem{li2019comparative}
Li, X., Wang, X., Xiao, G.: A comparative study of rank aggregation methods for
  partial and top ranked lists in genomic applications. Briefings in
  bioinformatics  \textbf{20}(1),  178--189 (2019)

\bibitem{Mallows1957}
Mallows, C.: Non null ranking models {I}. Biometrika  \textbf{44},  114--130
  (1957)

\bibitem{mosteller1951experimental}
Mosteller, F., Nogee, P.: An experimental measurement of utility. Journal of
  Political Economy  \textbf{59}(5),  371--404 (1951)

\bibitem{RankCentrality}
Negahban, S., Oh, S., Shah, D.: Iterative ranking from pair-wise comparisons.
  In: Advances in Neural Information Processing Systems 25. pp. 2474--2482
  (2012)

\bibitem{Negahban_RankCentrality_2017}
Negahban, S., Oh, S., Shah, D.: Rank centrality: Ranking from pairwise
  comparisons. Operations Research  \textbf{65}(1),  266--287 (2017).
  \doi{10.1287/opre.2016.1534}

\bibitem{niranjan2017inductive}
Niranjan, U., Rajkumar, A.: Inductive pairwise ranking: going beyond the n log
  (n) barrier. In: Thirty-First AAAI Conference on Artificial Intelligence
  (2017)

\bibitem{page1999pagerank}
Page, L., Brin, S., Motwani, R., Winograd, T.: The {PageRank} citation ranking:
  Bringing order to the {W}eb. In: Proceedings of the 7th International World
  Wide Web Conference. pp. 161--172 (1998)

\bibitem{shuman2013emerging}
Shuman, D.I., Narang, S.K., Frossard, P., Ortega, A., Vandergheynst, P.: The
  emerging field of signal processing on graphs: Extending high-dimensional
  data analysis to networks and other irregular domains. IEEE signal processing
  magazine  \textbf{30}(3),  83--98 (2013)

\bibitem{sync}
Singer, A.: Angular synchronization by eigenvectors and semidefinite
  programming. Appl. Comput. Harmon. Anal.  \textbf{30}(1),  20--36 (2011)

\bibitem{springall1973response}
Springall, A.: Response surface fitting using a generalization of the
  bradley-terry paired comparison model. Journal of the Royal Statistical
  Society: Series C (Applied Statistics)  \textbf{22}(1),  59--68 (1973)

\bibitem{stuart2006multiple}
Stuart-Fox, D.M., Firth, D., Moussalli, A., Whiting, M.J.: Multiple signals in
  chameleon contests: designing and analysing animal contests as a tournament.
  Animal Behaviour  \textbf{71}(6),  1263--1271 (2006)

\bibitem{thurstone1927law.}
Thurstone, L.: A law of comparative judgement. Psychological Review
  \textbf{34},  278--286 (1927)

\bibitem{vigna2016spectral}
Vigna, S.: Spectral ranking. Network Science  \textbf{4}(4),  433--445 (2016)

\bibitem{WM09}
Whiting, M.J., Webb, J.K., Keogh, J.S.: Flat lizard female mimics use sexual
  deception in visual but not chemical signals. Proceedings of the Royal
  Society B: Biological Sciences  \textbf{276}(1662),  1585--1591 (2009)

\bibitem{williams2006gaussian}
Williams, C.K., Rasmussen, C.E.: Gaussian Processes for Machine Learning. MIT
  Press (2006)

\end{thebibliography}
\end{document}